\documentclass[a4paper,fleqn]{cas-sc}

\usepackage[authoryear,longnamesfirst]{natbib}

\def\tsc#1{\csdef{#1}{\textsc{\lowercase{#1}}\xspace}}
\tsc{WGM}
\tsc{QE}
\tsc{EP}
\tsc{PMS}
\tsc{BEC}
\tsc{DE}
\usepackage{amsmath,amsthm}
\usepackage{multirow}
\usepackage{adjustbox}
\usepackage{graphicx}
\usepackage{booktabs} 
\usepackage{subfigure}
\usepackage{subcaption}
\usepackage{float}

\begin{document}
\let\WriteBookmarks\relax
\def\floatpagepagefraction{1}
\def\textpagefraction{.001}

\shorttitle{}    

\shortauthors{}  

\title [mode = title]{MSBATN: Multi-Stage Boundary-Aware Transformer Network for Action Segmentation in Untrimmed Surgical Videos}   

\tnotemark[1] 

\tnotetext[1]{} 

%

\author[1]{Rezowan Shuvo} [orcid=0000-0002-4186-2817,]

\cormark[1]

\fnmark[1]

\ead{m.shuvo@rgu.ac.uk}

\ead[url]{}

\credit{Conceptualization, Methodology, Writing- Original draft, review \& editing,, Visualization, }

\affiliation[1]{organization={School of Computing Engineering Technology, Robert Gordon University},
    addressline={Garthdee House}, 
    city={Aberdeen},
    postcode={AB10 7AQ}, 
    country={United Kingdom}}
    
\author[1]{M S Mekala}[orcid=0000-0003-2472-0291,]

\ead{ms.mekala@rgu.ac.uk}
\credit{Writing – Original draft, review \& editing, Validation, Supervision }

\author[1]{Eyad Elyan}[orcid=0000-0002-8342-9026,]
\ead{e.elyan@rgu.ac.uk}
\credit{Writing – review \& editing, Validation, Supervision, Resources}









\begin{abstract}
Understanding actions within surgical workflows is critical for evaluating post-operative outcomes and enhancing surgical training and efficiency. Capturing and analyzing long sequences of actions in surgical settings is challenging due to the inherent variability in individual surgeon approaches, which are shaped by their expertise and preferences. This variability complicates the identification and segmentation of distinct actions with ambiguous boundary start and end points. The traditional models, such as MS-TCN, which rely on large receptive fields, that causes over-segmentation, or under-segmentation, where distinct actions are incorrectly aligned. To address these challenges, we propose the Multi-Stage Boundary-Aware Transformer Network (MSBATN) with hierarchical sliding window attention to improve action segmentation. Our approach effectively manages the complexity of varying action durations and subtle transitions by accurately identifying start and end action boundaries in untrimmed surgical videos. MSBATN introduces a novel unified loss function that optimises action classification and boundary detection as interconnected tasks. Unlike conventional binary boundary detection methods, our innovative boundary weighing mechanism leverages contextual information to precisely identify action boundaries. Extensive experiments on three challenging surgical datasets demonstrate that MSBATN achieves state-of-the-art performance, with superior F1 scores at 25\% and 50\% thresholds and competitive results across other metrics.

\end{abstract}




\maketitle


\section{Introduction}
\label{introduction}
Surgical action segmentation involves analyzing instruments, organs, fluids, and smoke to identify and categorise distinct actions or phases in complex surgical videos \cite{mascagni2022computer}. This process is essential to improve surgical training by providing detailed feedback to trainees and advancing the performance of autonomous surgical robots. These robots can understand and perform intricate tasks, thus improving workflow efficiency and patient safety through real-time monitoring and optimisation of procedures. Recent advances in this field have led to the development of sophisticated architectures tailored to the unique challenges of surgical environments, such as Gazebo and the ROS2 software development kit (SDK). These innovations support better decision-making during operations and drive progress in surgical robotics and education \cite{kiyasseh2023vision}.

\begin{figure}
    \centering    
    \includegraphics[width=.6\textwidth]{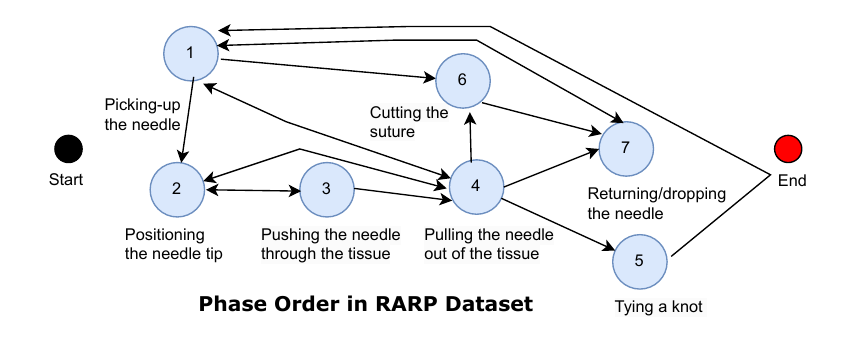}

    \caption{ The action-transition graph for SAR-RARP50 features nodes 1 to 7, representing canonical suturing actions. A typical sequence is: Pick up the needle (1) → Position the tip (2) → Push through the tissue (3) → Pull out (4). After this sequence, the surgeon can either return or drop the needle (7), tie a knot (5), cut the suture (6), or skip the last two steps. This graph allows for loops and re-entries, so it doesn’t follow a single linear path from start to finish.}
    \label{fig:action_order}
\end{figure}
 
Surgical videos are typically annotated manually, with each frame labeled according to the surgical phase, sub-phase, or specific actions/gestures, as well as the instruments present in that frame \cite{psychogyios2023sar, twinanda2016endonet}. Recent datasets have introduced action triplets \cite{nwoye2022rendezvous}, which include information about the action, instruments, and organs for frame-wise action recognition and segmentation of instruments or organs \cite{nwoye2023cholectriplet2021, chadebecq2023artificial}. Most of these datasets focus primarily on spatial information, while providing minimal temporal information.
 
There is a standard temporal hierarchy that surgeons follow during procedures. The highest-level temporal component of a surgical operation or intraoperative activity is referred to as a "phase," which encompasses a series of related activities. Each phase consists of several "steps," which are the specific actions taken to achieve the goals of that phase. Each step represents a procedure-specific segment aimed at addressing the patient's condition, while the "task" is a sub-component of a step that involves a series of actions \cite{meireles2021sages}. 
However, the definitions of temporal evolution hierarchies can be subjective and depend on the surgeon's experience, which is why they are rarely disclosed alongside the videos. This lack of structure, coupled with visual similarities between actions and fuzzy boundaries, often leads to ambiguous transitions (Figure~\ref{fig:action_order}). For example, in SAR-RARP50, the action-transition graph is highly cyclic: any of the seven suturing gestures can be repeated or skipped, resulting in frequent re-entries and out-of-order sequences.

Deep learning pipelines for surgical video analysis have advanced significantly, evolving from recurrent networks to temporal convolutions, and more recently, to Transformers \cite{farha2019ms,li2020ms,czempiel2020tecno,yi2022not,yang2024surgformer,zhang2022actionformer,ding2022exploring,nwoye2022rendezvous,shi2023tridet}. Early systems, such as EndoNet \cite{twinanda2016endonet} and EndoLSTM \cite{twinanda2017vision}, utilised CNN backbones for frame features and RNN modules for temporal smoothing, achieving the first joint modeling of tool presence and phase recognition. Subsequent variants incorporated correlation losses, bidirectional LSTMs, or multitask heads to improve accuracy \cite{mondal2019multitask,jin2020multi}. However, RNN-based designs often struggle to retain dependencies over minutes—to hours—long procedure.

To expand the temporal receptive field without encountering the vanishing gradient problems associated with RNNs, researchers have shifted to Temporal Convolutional Networks (TCNs)) \cite{farha2019ms,li2020ms,czempiel2020tecno,yi2022not}. The MS-TCN family employs dilated 1-D kernels that efficiently aggregate historical information while maintaining a low inference latency \cite{farha2019ms,li2020ms}. Recently, single-stage Transformers like Actionformer \cite{zhang2022actionformer} and TriDet \cite{shi2023tridet} have provided global self-attention over short clips, excelling in fine-grained gesture localisation. However, these single-pass models were designed for brief human actions; their fixed window makes them unsuitable to capture the hierarchical nature of surgical workflows.
Multi-stage architectures address this limitation by incorporating one or more refinement passes to correct over- or under-segmentation produced in the first stage. TeCNO \cite{czempiel2020tecno} chains several TCN blocks, each trained to refine the output of its predecessor, while MTMS-TCN jointly learns phases and steps to leverage shared cues \cite{farha2019ms,li2020ms,ramesh2021multi}. Other research has integrated kinematic sensor streams or attention mechanisms to improve temporal reasoning \cite{van2022gesture,goldbraikh2023kinematic}, and segment-level Transformers have refined ambiguous boundaries by analysing high-level proposals \cite{ding2022exploring,jin2022exploring}. Despite these advancements, refinement modules often work with hard predictions without an explicit model of surgical semantics, which can lead to persistent systematic errors, particularly when actions visually resemble one another.

\begin{figure}
    \centering    \includegraphics[width=0.76\textwidth]{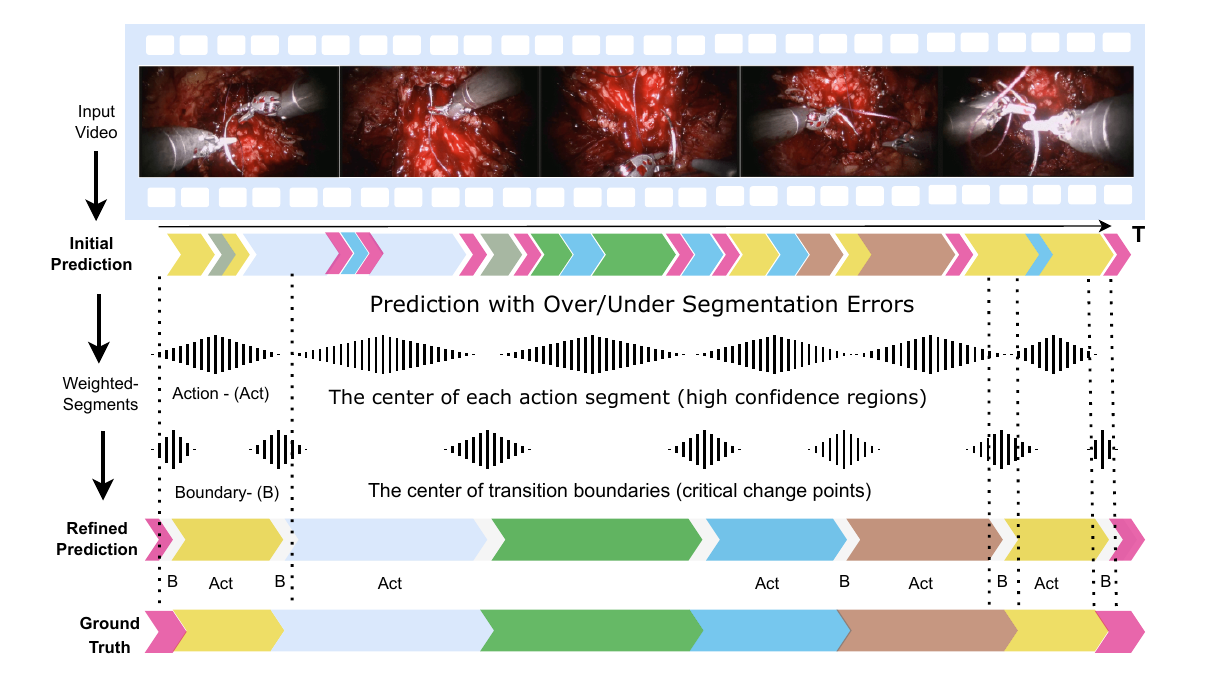}
    \caption{
    Multi-stage Boundary-Aware Refinement Pipeline: The initial prediction with typical over- and under-segmentation errors (segments are split too finely, merged, or slightly shifted) misleads action order. Our approach resolves this by applying a Gaussian centre-weighting scheme to Action and Boundary segments to minimise the error. The refined prediction aligns with the annotated ground truth, and boundaries without re-processing the video.}
    \label{fig:overview_framework}
\end{figure}

Furthermore, these approaches overlook the finer details that are essential for understanding the intricate relationships between spatial and temporal features \cite{kiyasseh2023vision}, resulting in over and under-segmentation errors. The \textit{over-segmentation} occurs when a segment is incorrectly divided into multiple action segments, leading to the detection of more actions than actually exist. These approaches are overly sensitive to minor variations in the input data, leading them to interpret slight changes as new action boundaries \cite{ishikawa2021alleviating}. On the other hand, \textit{under-segmentation} happens when distinct actions are mistakenly merged into a single segment. \textit{This typically arises due to a lack of sufficient temporal resolution or the inability to recognise subtle transitions between different actions.} So, the model overlooks the actual action boundaries, resulting in fewer detected action segments than actually exist.

The continuous and smooth transitions characteristic of surgical activities are often ignored due to treating actions and boundaries as discrete events without accounting for their gradual shifts between actions \cite{fard2017soft,ishikawa2021alleviating}. We noticed that the self-attention mechanism has difficulty focusing on important local areas, particularly at the initial layers. Additionally, the encoder-decoder architectures do not effectively refine predictions based on the temporal relationships between actions \cite{yi2021asformer}. Our solution (Fig. \ref{fig:overview_framework}) tackles these problems in two steps: firstly, a backbone network predicts initial labels for actions and their boundaries. Secondly, a lightweight post-processing module corrects over- and under-segmentation by focusing exclusively on the start–end transitions. Because it operates only on predicted segments, it adds negligible computational cost to the main model. Training utilises a duration-aware loss weighting that adjusts each segment’s contribution inversely based on its duration, ensuring that short yet clinically significant actions are not overshadowed by longer ones.

We propose a framework that extracts spatial features with DinoV2 by discarding irrelevant large patch tokens with register tokens \cite{oquab2023dinov2}. For temporal modelling, we present an encoder-decoder architecture that is built based on the combination of dilated TCN, unified hierarchical and sliding window attention, which captures fine-to-coarse context. The hierarchical representation with dynamic sliding window attention is used to handle long input sequences that organise the attention layers hierarchically to first focus on local relationships and then gradually capture global dependencies. We design a unified loss function by combining the loss functions of individual tasks into a single action segmentation to achieve effective boundary detection. Our approach introduces the following key innovations:

\begin{itemize}
    \item We present a multi-stage TCN-Transformer architecture with dilated acausal convolution to widen the receptive field that predicts and refines both actions and boundaries. 
    \item We introduce hierarchical sliding window attention mechanisms (HSWA) to resolve ambiguities among action orders from different temporal resolutions. It links high-level action segment structures with fine-grained frame-wise details, enabling the model to better understand long-term temporal dependencies and immediate frame-level changes.
    \item  We propose a boundary-aware loss function that gradually measures cosine similarity from centre to boundaries inside action segments that assist in minimising over- or under-segmentation error.

\end{itemize}

The remainder of this article is structured as follows: Section \ref{sec2} focuses on related work. Section \ref{sec3} describes the proposed framework. The model performance is discussed based on experimental evaluation results in Section \ref{sec4}, and the article concludes with Section \ref{sec5}.

\section{Related Work}\label{sec2}

\subsection{Action Recognition and Segmentation}
For the surgical domain, EndoNet \cite{twinanda2016endonet} was developed for phase recognition in cholecystectomy videos by learning inherent visual features through a CNN architecture. Recognising the importance of tool information in phase recognition, EndoNet adopted a multi-task framework that incorporates tool presence detection into the feature learning process. An extension of AlexNet was designed with dual objectives: one part focused on \textit{detecting the presence of surgical tools}, while the other was responsible for \textit{recognising seven distinct surgical phases}. To capture temporal dependencies, EndoNet integrated a Hierarchical Hidden Markov Model (HMM), enabling the modelling of temporal transitions between phases. The architecture is further improved by LSTM \cite{twinanda2017vision} and ResNet \cite{twinanda2018rsdnet}. However, these methods struggle with accurately identifying transition frames between action phases. To overcome this limitation, a sequence-to-sequence model with an attention mechanism was proposed \cite{namazi2019attention,yi2021asformer}, which accommodates variable-length inputs and outputs, making it well-suited for detecting overlapping surgical tasks.

Temporal Action Detection (TAD) methodologies are mainly classified into two-stage and one-stage-based methods. The two-stage approach, inspired by object detection frameworks like Faster R-CNN, involves an initial phase of generating action proposals, followed by subsequent stages of classification and refinement \cite{chao2018rethinking,keisham2022online}. Although this method often achieves higher accuracy due to its detailed localisation process, \textit{it tends to be computationally intensive, posing challenges for real-time applications} \cite{yi2022not}. On the other hand, the one-stage approach consolidates action proposal generation and classification in a single framework, which offers computational efficiency for real-time scenarios. Models like ActionFormer \cite{zhang2022actionformer}, and Surgformer \cite{yang2024surgformer} are transformer-based architectures that have achieved enhanced precision of action localisation while maintaining computational efficiency. These models capture long-range dependencies and contextual information that helps improve the accuracy of action segmentation and detection. However, \textit{one-stage methods have historically faced difficulties in precisely localising action boundaries in long untrimmed videos due to their small receptive fields, which is crucial for optimal performance in TAD}.  

\subsection{Boundary Detection}

Identifying action boundaries within untrimmed videos poses unique challenges due to the inherent ambiguity in determining precise start and end points of actions. Initial action boundary detection approaches used probabilistic models to capture temporal dynamics. For example, a spatiotemporal Gaussian Mixture Model (GMM) with Variational Bayesian (VB) \cite{loukas2016shot} inference was employed to dynamically estimate the number of components in the GMM, which eliminates the need to predefine the number of clusters. VB inference approximated the full posterior distribution, which improved model flexibility and reduced redundant components. Label assignment and tracking were accomplished using Kullback-Leibler (KL) distance, ensuring consistent tracking across video clips. Unlike rigid boundary detection methods, the Soft-Boundary Unsupervised Gesture Segmentation (Soft-UGS) approach \cite{fard2017soft} was proposed as a soft-boundary clustering technique that allowed for gradual transitions between actions. This method effectively utilised kinematic trajectory data, Dynamic Time Warping (DTW), and probabilistic clustering to segment surgical gestures in real-time applications. Although these probabilistic models provided a foundation for more advanced techniques, they faced limitations in precision, especially in detecting abrupt transitions in video content.

In contrast, the emergence of deep learning introduced new techniques for addressing the temporal and spatial complexities of action boundary detection. For the surgical domain, sequence models were applied to map variable-length input sequences to output sequences while incorporating attention mechanisms for more accurate boundary localisation. A double attention mechanism(such as \cite{namazi2019attention}) was introduced to focus specifically on the beginning and end of each surgical phase, offering a refined approach to boundary detection in complex tasks like surgery.

The above methods for temporal segmentation face the issue of over-segmentation because they segment actions into too many sub-components, leading to inaccurate predictions. The Action Segment Refinement Framework (ASRF) \cite{ishikawa2021alleviating} was introduced to address this problem by decoupling frame-wise classification from boundary regression. ASRF refines action segmentation through two branches: one for action predictions and another for boundary regression. By focusing specifically on predicted action boundaries, ASRF achieved overall action localisation accuracy.
ASRF* \cite{yi2021asformer} improved the performance with an attention mechanism.

Recent methods, such as TriDet \cite{shi2023tridet} and Elastic Moment Bounding (EMB) \cite{huang2022video}, have further extended the capabilities of one-stage detection for real-time applications. TriDet introduced the Trident-head detection mechanism, which models action boundaries by estimating relative probability distributions instead of directly predicting offsets. \textit{This innovation significantly improved boundary localisation accuracy while reducing computational overhead.}
EMB tackles uncertainties in temporal boundaries by transforming rigid and manually labelled endpoints into an elastic set of boundaries; however, it does tend to generate a high number of false positives.

Building upon these advancements, our action boundary network employs a voting mechanism to improve segmentation by accurately finding the start and end of an action. By utilising the predicted action boundaries, our framework refines frame-wise predictions, ultimately improving the performance of action segmentation.

\section{Proposed Framework}\label{sec3}
The action segmentation task involves predicting the action category for each frame in a video. Formally, given a sequence of \( T \)-frame features \(
X = [\mathbf{x}_1, \mathbf{x}_2, \ldots, \mathbf{x}_T] \in \mathbb{R}^{T \times d}, \quad \text{where } \mathbf{x}_T \in \mathbb{R}^d\) represents the \( d \)-dimensional feature of the \( t \)-th frame. The objective is to predict a sequence of actions \(\hat{Y} = [\hat{y}_1, \hat{y}_2, \ldots, \hat{y}_T], \quad \text{where } \hat{y}_T \in C.
\) The set \( C = \{0, 1, 2, \ldots, c\} \) includes all action categories in the dataset. The frame-wise ground truth labels are \(
Y = [y_1, y_2, \ldots, y_T],\) accompanied by segment information
\(S = \{(y_{s_1}, y_{e_1}, a_{c_1}), \ldots, (y_{s_n}, y_{e_n}, a_{c_n})\},
\) where \( y_{s} \) and \( y_{e} \) denote the start and end of a segment, \( a_c \) represents the action class label, and \( n \) is the number of segments, including background segments. \textit{The goal is to accurately predict action transitions and segment boundaries.} Since action transitions are continuous and boundaries are often subjective, considering this 5\% of the total segment length is labeled as a transition buffer zone at the start and end of each segment.

\begin{figure*}
    \centering    \includegraphics[width=1\textwidth]{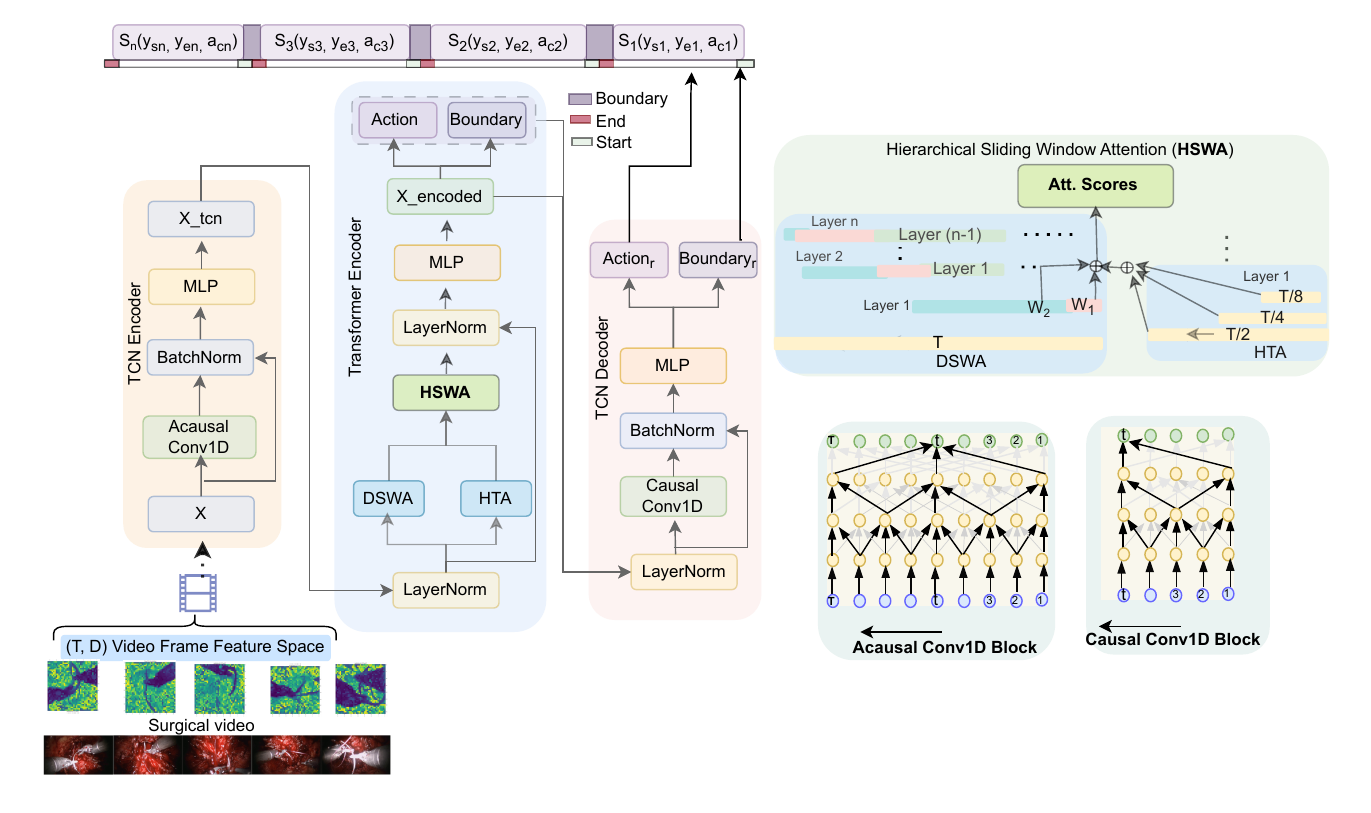}
    \caption{ Proposed Multi-Stage Boundary-Aware Transformer Network (MSBATN): Input video features $(X)$ are first processed by TCN encoder consisting Acausal Conv1D (\textcolor{red}{see right bottom}) for temporal feature extraction to produce $X_{tcn}$. These features are then fed into the Transformer Encoder, consisting of Hierarchical Sliding-Window Attention (HSWA) (\textcolor{red}{see right top}), to generate $X_{encoded}$. The $X_{encoded}$ features are expanded to the original temporal resolution to separate Action and Boundary segmentation heads. Finally, a TCN Decoder refines these initial predictions to output the final action segments and boundaries.}
    \label{fig:our_architecture}
\end{figure*}

Our proposed framework, the Multi-Stage Boundary-Aware Transformer Network (MSBATN), is illustrated in Figure \ref{fig:our_architecture}. MSBATN is designed to improve action segmentation in untrimmed surgical videos by accurately identifying action boundaries, even in lengthy and complex sequences. It achieves this through three main components: robust frame-wise feature extraction, an encoder-decoder architecture based on TCN-Transformer-TCN that incorporates Hierarchical Sliding Window Attention (HSWA) and specialised acausal/causal convolutions, and a novel boundary-aware loss function.

\subsection{Frame-Wise Feature Extraction
}

\begin{figure}
    \centering    \includegraphics[width=.6\textwidth]{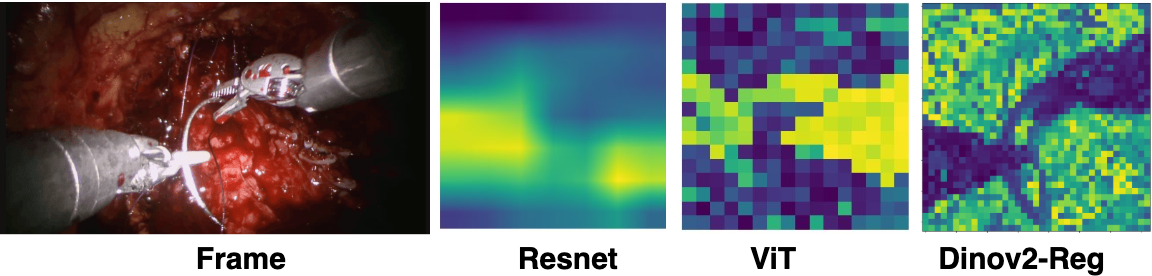}
    \caption{Comparison of frame feature extractors.}
    \label{fig:compare_dino}
\end{figure}

The pipeline begins with an input surgical video from which individual frames are extracted. For MSBATN, we specifically use Dinov2-Reg, a feature extractor that improves upon standard ViTs. In standard ViTs, images are divided into patches represented by tokens, but "outlier tokens" can introduce errors due to unusually high values. Dinov2-Reg addresses this by adding register tokens as dedicated placeholders, enhancing feature robustness and reducing reliance on outlier tokens. As shown in Figure~\ref{fig:compare_dino}, Dinov2-Reg \cite{oquab2023dinov2} captures complex spatial features more effectively in a self-supervised manner compared to standard ViT and ResNet methods.

The Dinov2-Reg extractor converts each frame into a high-dimensional patch-based feature representation (e.g., resulting in a ($H' \times W' \times C_{dino}$) tensor. We use (256, 1536), where 256 denotes the $16 \times 16$ patch dimension, and 1536 is the feature dimension. To create a fixed-size feature vector per frame suitable for our temporal model, we project these features. Specifically, we apply a 2D convolutional layer (or an equivalent linear projection on flattened patch features) to transform the DINOv2 output for each frame into a (d)-dimensional vector (d=2048). This results in an initial feature sequence X with dimensions (T, D), where T is the sequence length. This step standardises the frame features and prepares them for the subsequent MSBATN architecture.

\subsection{Multi-Stage Boundary-Aware Transformer Network (MSBATN)}

The core of our framework is the MSBATN, a Transformer-inspired encoder-decoder architecture designed for effective temporal feature learning and segmentation (see Figure \ref{fig:our_architecture}). It includes an initial TCN-based feature processor, a Transformer-based encoder for deeper contextual understanding, and TCN-based decoders for refinement. Notably, MSBATN does not need explicit positional encodings, as its design naturally captures sequence order and relationships through temporal convolutions and attention mechanisms.
\subsubsection{Encoder-Decoder}

The encoder path begins by processing the frame-wise features  $X = (B, D_{in}, T_{orig})$, where B is batch size, $D_{in}$ is the input feature dimension, and $T$ is the original sequence length. The input features first pass through an initial TCN stage. It projects the input dimension (2048) to the model's working dimension (256). It employs dilated 1D Acausal TCN where the kernel k is centred on the current frame, covering T-K T, T+K. In Causal TCN, the kernel K shifts from right to left, that is, T-K to T thus never peeks into the future, and therefore safe for real-time or streaming inference (Figure~\ref{fig:our_architecture}). When the sequence is too long we applied stride greater than 1 to reduce the sequence length $T$ to $T_{reduced}$ resulting in $X_{tcn} = (B,256,T_{reduced})$. This makes subsequent attention computations more efficient. The dilated convolutions allow the network to build a hierarchical receptive field, capturing local temporal patterns and dependencies over extended sequences effectively. These TCN blocks can access both past and local future context within their receptive field during training, enabling robust local feature extraction.

$X_{tcn}$ is then processed by a stack of encoder blocks. Each block aims to capture more complex and longer-range temporal dependencies. Each encoder block consists of Dual Sliding Window Attention (DSWA), Hierarchical Temporal Attention (HTA) and Feed-Forward Network (MLP) with Layer Normalisation and Residual Connections. Lastly the initial predictions from the encoder are refined by the decoder.  
Instead of acausal convolution in the encoder TCN, the decoder uses causal convolution to avoid information about the future. 
The network is comprised of one encoder and three decoders, with each encoder and decoder containing ten TCN blocks. We set the feature dimension in the encoder and decoder layers to 256. Additionally, a temporal dropout layer is applied to the input feature of the encoder, which randomly drops the entire feature channel with a dropout rate of 0.3 to mitigate the over-fitting issue.

\subsubsection{Hierarchical Sliding Window Attention}
We present a Hierarchical Sliding Window Attention (HSWA) module that effectively captures temporal dependencies by combining dynamic dual-sliding window attention (DSWA) with hierarchical temporal attention for global context.
In DSWA, instead of dense self-attention, we use a sparse attention approach inspired by Longformer \cite{beltagy2020longformer}. For each query token (frame representation), attention is computed over two distinct local windows. Across the stack of encoder layers, one set of windows progressively expands (e.g., from a one-sided window of 16 up to 256, for an effective receptive field of 32 to 512), while the other set of windows symmetrically shrinks. This dynamic dual-view allows the model to capture dependencies at varying scales throughout the network depth.
Within each sliding window, attention is dilated (e.g., with rates 0, 1, 2, \ldots) in deeper layers. The dilation rates start from 0, meaning some windows, especially in earlier layers or specific attention heads, operate without dilation, focusing squarely on immediate local context. This allows the window to cover a larger receptive field without increasing the number of attended tokens. This dynamically sized windows, ensures that smaller, more focused windows are always present to capture fine-grained local details, even as other windows expand or become more dilated to see broader context. To complement the local/sparse sliding windows and capture broader context, we incorporate a HTA at multiple temporal scales. The sequence length $T_{reduced}$ is conceptually downsampled e.g., $T_s = T_{\text{reduced}} / 2^s$, for scales $ (s = 0, 1, 2, \ldots)$. At each scale $(T_s)$, sparse attention is computed. The number of scales can be determined by $\log_2(T_{\text{reduced}}/S_{\text{avg}})$, where $(S_{\text{avg}})$ might relate to an average segment length or a minimum sensible sequence length for global attention (e.g., until $(T_s \geq 512)$ as mentioned, though this was for original T). The attention scores from these different scales are then aggregated to form a "global" contextual representation. Let $(e_{ij}^s) $ be the attention score between token $(x_i)$ and $(x_j)$ within its neighborhood $(\mathcal{N}_{i^s})$ at scale (s). The aggregated attention score considers information across scales: 
\begin{equation} \alpha{ij} = \frac{\exp\left(\sum_s w_s \cdot e_{ij}^s\right)}{\sum_{k \in \bigcup_s \mathcal{N}i^s} \exp\left(\sum_s w_s \cdot e{ik}^s\right)} 
\end{equation} where$ (w_s)$ could be learnable weights or fixed, and the neighborhood $(\mathcal{N}_i)$ spans relevant tokens across scales.
We believe these design elements, ensure that the HSWA captures sufficient local details while effectively balancing the learning of both local and global information. The ablation studies presented in Section 4.3.1, especially regarding the performance of ACHDSWO (Acausal Hierarchical Dual Sliding Window with Overlap), further validate the effectiveness of this combined approach.

\subsection{Boundary-aware Segmentation Loss }
\begin{figure}
    \centering    \includegraphics[width=.7\textwidth]{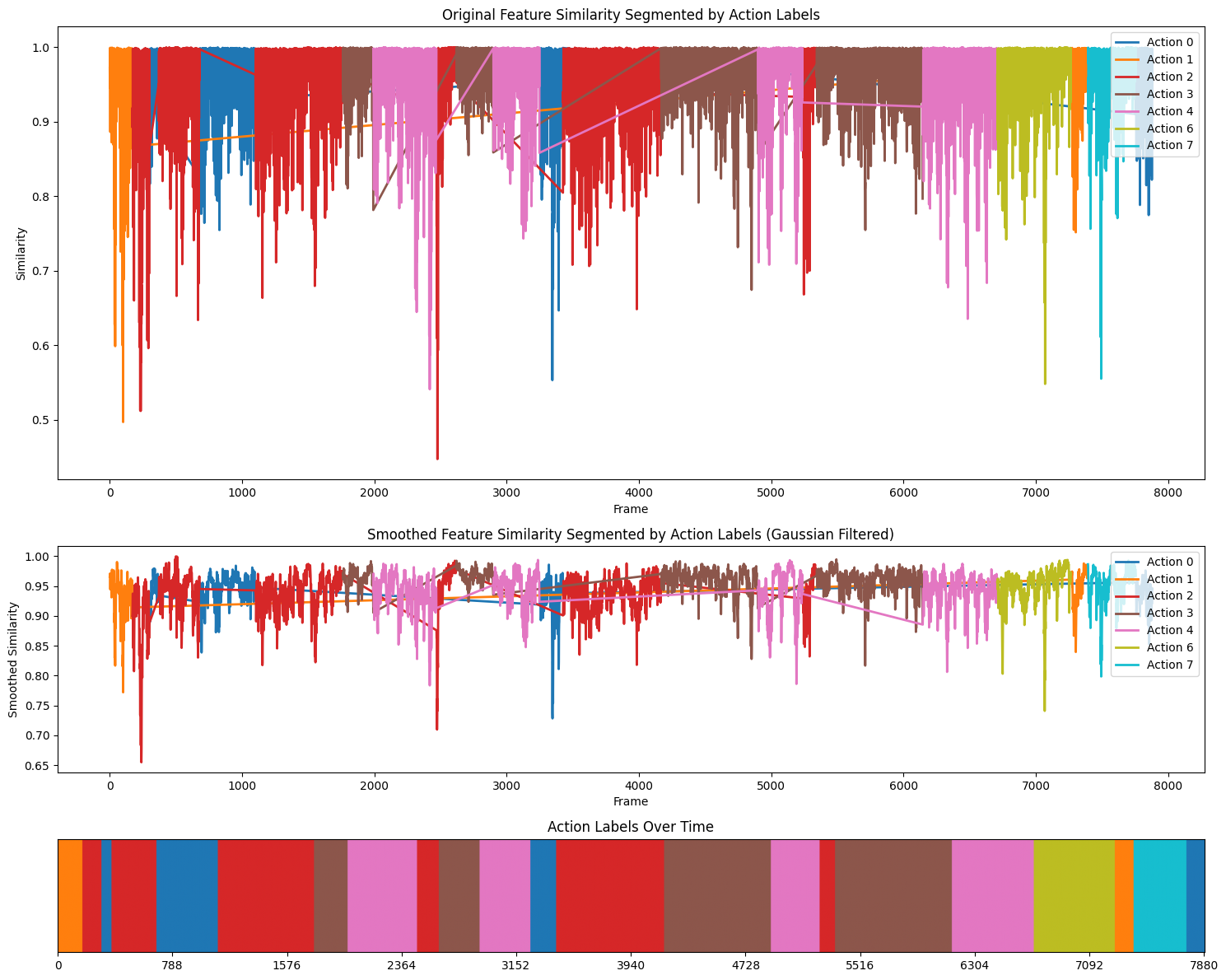}
    \caption{Segment-wise Gaussian-cosine similarity scores for SAR-RARP}
    \label{fig:gauss_sim}
\end{figure}
In temporal action segmentation, accurately capturing action transitions and segment-level dynamics across varying time scales is crucial. Surgical videos are typically long, and the duration of the segment fluctuates significantly, as shown in Table \ref{tab:merged_table_no_count}.  To address this, we propose a \textit{Boundary-Aware Segmentation Loss ($\mathcal{L}_{\text{temporal}}$ )}, a unified objective function that integrates multiple specialised loss components. To ensure precise action classification at each timestep and handle class imbalances prevalent in surgical datasets, we employ Focal Loss $\mathcal{L}_{\text{class}}$. 
To enhance the overlap between predicted and ground truth segments, we utilise Dice Loss derived from the Dice Coefficient:
\begin{equation}
  \text{Dice Coefficient} = \frac{2 \times |A \cap B|}{|A| + |B|}  
\end{equation}
where \(A\) and \(B\) represent the predicted and ground truth sets, respectively. The Dice Loss is then formulated as:
\begin{equation}
 \mathcal{L}_{\text{dice}} = 1 - \text{Dice Coefficient}   
\end{equation}
This formulation emphasises the overlap between predictions and ground truth, making it sensitive to both false positives and false negatives.
While Focal and Dice losses address classification and overlap, they do not explicitly model the temporal smoothness of feature representations within an action segment. As illustrated in Figure \ref{fig:gauss_sim}, raw frame-to-frame feature similarities can be ambiguous. To encourage temporal consistency and smooth transitions, $\mathcal{L}_{\text{sim}}$ penalises deviations in feature representations $\mathbf{f}_t$ between consecutive frames. \textit{This penalty is weighted by a Gaussian function \(G(t, t+1)\) which is designed to peak at the temporal centre of each ground truth action segment, diminishing towards its start and end points.} This ensures that the strongest enforcement of feature similarity occurs within the core of an action, while allowing for more natural feature evolution near the transitions to adjacent actions, promoting coherent feature sequences.

\begin{equation}
 \mathcal{L}_{\text{sim}} = \sum_{t=1}^{T-1} G(t, t+1) \left(1 - \cos(\mathbf{f}_t, \mathbf{f}_{t+1})\right); G(t) = \exp\left(-\frac{(t - c)^2}{2\sigma^2}\right) 
   \end{equation}

Where  $c$ is the temporal centre of the segment, $\mathbf{f}_t$ is the feature vector of frame $t$, and $\sigma$ is a hyperparameter controlling the width of the Gaussian weighting.
This formulation ensures that the penalty for feature dissimilarity between consecutive frames ($\mathbf{f}_t, \mathbf{f}_{t+1}$) is highest when those frames (referenced by index $t$) are near the centre ($c$) of their action segment and lowest when they are near the boundaries. This aligns perfectly with the goal of encouraging strong feature similarity at the "core of an action" while allowing for more natural feature evolution near the transitions.

To further refine boundary precision, we incorporate a Gaussian Similarity Truncated Mean Squared Error (MSE) loss. This loss focuses on boundary regions, applying a Gaussian weight to the MSE, which emphasises errors near segment boundaries:
\begin{equation}
 \mathcal{L}_{\text{boundary}} = \sum_{t=1}^{T} G_{\text{boundary}}(t) \cdot \min\left((\hat{b}_t - b_t)^2, \tau\right)   
\end{equation}
Where \(G_{\text{boundary}}(t)\) is \textit{a Gaussian weight which is designed to have its highest values precisely at the annotated start and end frames of action segments (i.e., at the actual boundary points between differing actions), effectively creating high-importance zones for the loss calculation at these critical transitions.}  \(\hat{b}_t\) and \(b_t\) are the predicted and ground truth boundary positions at frame \(t\), and \(\tau\) is a truncation threshold to limit the influence of large errors. By integrating these loss components, the total temporal loss function is defined as:
\begin{equation}
 \mathcal{L}_{\text{temporal}} = \alpha \cdot \mathcal{L}_{\text{class}} + \beta \cdot \mathcal{L}_{\text{dice}} + \gamma \cdot \mathcal{L}_{\text{sim}} + \delta \cdot \mathcal{L}_{\text{boundary}}   
\end{equation}
Where \(\alpha\), \(\beta\), \(\gamma\), and \(\delta\) = (1.0,0.2,0.5,0.5) were empirically determined on our validation set. This weighting factors balance the contributions of each loss term. This comprehensive loss function enables the model to effectively capture class imbalances, segment overlaps, temporal dependencies, and boundary precision, leading to improved performance in temporal action segmentation tasks.

\section{Experiments and Analysis}\label{sec4}

\begin{table}[]
\centering
\caption{Class  Duration in SAR-RARP and Cholec80 Datasets}
\label{tab:merged_table_no_count}
\resizebox{.605\textwidth}{!}{
\begin{tabular}{l l l l }
\hline
\textbf{Dataset} & \textbf{ID} & \textbf{Classes/Phases} & \textbf{Mean$\pm$Std (s)} \\
\hline
\multirow{8}{*}{\rotatebox{90}{SAR-RARP}} 
 & 0 & Background Classes & 12.4 $\pm$ 17.2 \\
 & 1 & Picking-up the needle & 3.84 $\pm$ 2.66 \\
 & 2 & Positioning the needle tip & 6.83 $\pm$ 5.65 \\
 & 3 & Pushing needle through tissue & 7.51 $\pm$ 3.79 \\
 & 4 & Pulling the needle out of tissue & 6.77 $\pm$ 3.73 \\
 & 5 & Tying a knot & 26 $\pm$ 7.65 \\
 & 6 & Cutting the suture & 6.89 $\pm$ 5.31 \\
 & 7 & Returning the needle & 6.91 $\pm$ 5.41 \\
\hline
\multirow{7}{*}{\rotatebox{90}{Cholec80}}
 & P1 & Preparation & 125 $\pm$ 95 \\
 & P2 & Calot triangle dissection & 954 $\pm$ 538 \\
 & P3 & Clipping and cutting & 168 $\pm$ 152 \\
 & P4 & Gallbladder dissection & 857 $\pm$ 551 \\
 & P5 & Gallbladder packaging & 98 $\pm$ 53 \\
 & P6 & Cleaning and coagulation & 178 $\pm$ 166 \\
 & P7 & Gallbladder retraction & 83 $\pm$ 56 \\
\hline
\end{tabular}
}
\end{table}

\begin{table*}[ht!]
\caption{Comparison with state-of-the-art models Across Datasets (SAR-RARP, Cholec80, JIGSAW)}
\label{tab:sotacompare}
\centering
\adjustbox{max width=1\textwidth}{%
\begin{tabular}{lllllll}
\toprule
\textbf{Dataset} & \textbf{Model} & \textbf{Acc} & \textbf{Edit} & \textbf{F1@10} & \textbf{F1@25} & \textbf{F1@50} \\ 
\midrule

\multirow{5}{*}{\rotatebox{90}{\textbf{SAR-RARP}}} 
    & MS-TCN\cite{li2020ms}        & 69.0        & -           & 70.7        & –              & –              \\ 
    & ActionCLIP\cite{psychogyios2023sar}    & 80.4          & –              & 80.6           & –              & –              \\ 
    & MA-TCN\cite{van2022gesture}            & 83.4          & 81.6           & 81.7           & –              & –              \\ 
    & ASRF\cite{ishikawa2021alleviating} & $87.5 \pm 3.7$ & $83.6 \pm 1.9$   & $80.1\pm 3.2$ & $70.8 \pm 6.2$ & $59.6 \pm 3.2$ \\

    & NETE\cite{yi2022not}                   & $87.8 \pm 3.6$        & $ 83.9 \pm 1.5$            & $78.7 \pm 3.6$           & $68.5 \pm 3.2 $             & $54.6 \pm 2.4$           \\ 
     & Asformer\cite{yi2021asformer}          & $\boldsymbol{88.8 \pm 5.4}$         &   $\boldsymbol{84.1 \pm 3.1}$        & $\boldsymbol{84.3 \pm 5.3}$           & $ 68.1\pm 3.4 $           & $59.4 \pm 3.1$             \\ 
    & MSBATN (Ours)                           & $88.7 \pm 3.4$ & \textbf{ $84.3 \pm 2.7$}          & $84.0 \pm 3.7$           & $\boldsymbol{72.4 \pm 2.6}$           & $ \boldsymbol{63.8 \pm  1.5} $        \\ 
\midrule

\multirow{6}{*}{\rotatebox{90}{\textbf{Cholec80}}} 
    & ResNet\cite{yi2022not}                 & 78.3          & 57.8           & 52.2           & –              & –              \\ 
    & PhaseLSTM\cite{neil2016phased}         & 80.7          & 61.2           & 62.4           & –              & –              \\ 
    & EndoNet\cite{twinanda2016endonet}      & 81.9          & –              & –              & –              & –              \\ 
    & TeCNO\cite{czempiel2020tecno}          & 88.6          & –              & –              & –              & –              \\ 
     & ASRF\cite{ishikawa2021alleviating}          & $82.5\pm4.5 $         &  $78.5 \pm 3.5$              & $84.3 \pm3.4$           & $70.6\pm5.7 $              & $60.1\pm3.4 $              \\ 
    
    & Asformer\cite{yi2021asformer}          & $82.8 \pm 4.7$         &  $74.5 \pm 6.7$           & $84.1 \pm2.4$           & $70.8\pm6.2 $            & $57.1\pm3.2 $              \\ 
    & NETE\cite{yi2022not}                   & $\boldsymbol{92.8\pm 5.0} $         & $\boldsymbol{78.7\pm 9.4}$             & $82.4 \pm3.4$          & $70.1\pm3.2 $             & $53.1\pm4.8 $            \\ 
    & MSBATN (Ours)                           & $89.4 \pm 2.1$ & $77.5 \pm 8.0$          & $\boldsymbol{84.4 \pm4.4}$           & $\boldsymbol{73.1\pm5.2}$           &$\boldsymbol{ 60.2\pm8.4}$   \\ 
\midrule

\multirow{3}{*}{\rotatebox{90}{\textbf{JIGSAW}}} 
    & K-TCN\cite{van2022gesture}             & 83.8          & 86.3           & 90.4           & –              & –              \\ 
    & MA-TCN\cite{van2022gesture}            & 86.8          & 90.1           & 93.6           & –              & –              \\ 
     & ASRF\cite{ishikawa2021alleviating}          & $87.8\pm4.5 $        & $89.0\pm2.5 $            & $80.4\pm 6.5$          & $70.1\pm 3.6$           & $69.8 \pm 2.5 $      \\ 
    
     & NETE\cite{yi2022not}                   & $90.8\pm2.1$          & $86.5\pm3.5 $              & $74.4\pm 7.8$          & $65.1\pm 3.7$           & $58.6 \pm 2.9 $      \\ 
     & Asformer\cite{yi2021asformer}          & $\boldsymbol{92.8\pm3.3}$          & $\boldsymbol{89.8\pm6.7 }$            & $83.5\pm 4.2$          & $70.1\pm 3.5$           & $65.4 \pm 1.8 $     \\ 
    & MSBATN (Ours)                           & $86.1 \pm 6.3$ & $89.5\pm6.5 $          & $\boldsymbol{84.4\pm 5.5}$           & $\boldsymbol{76.1\pm 3.1}$           & $\boldsymbol{70.7 \pm 2.8} $         \\ 
\bottomrule
\end{tabular}%
}
\end{table*}

\begin{figure}
    \centering    \includegraphics[width=.7\textwidth]{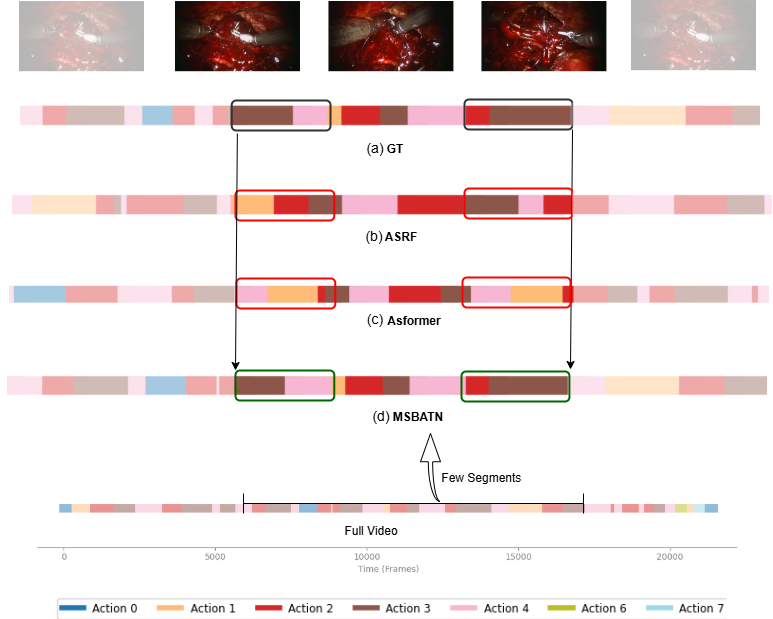}
    \caption{Qualitative action segmentation comparison. (a) Ground Truth (GT), where each colour label represents an action. The rectangular boxes explicitly highlight actions for detailed comparison of actions' start/end. (b) ASRF \cite{ishikawa2021alleviating} and (c) Asformer \cite{yi2021asformer} show deviations in action start/end points (highlighted in \textcolor{red}{red rectangular box}). (d) The proposed MSBATN model demonstrates better alignment with GT (highlighted in \textcolor{green}{green}). Segments are zoomed from the full video timeline below.}
    \label{fig:merged_temporal_comparison}
\end{figure}

The encoder ingests inputs of shape (T, 2048) and processes them through a dilated acausal TCN, effectively reducing the dimensionality to (T,256) while capturing temporal dependencies over varying scales. These embeddings are further refined through ten layers of multi-head sparse attention mechanisms (with 8 heads), incorporating advanced techniques such as sliding window attention for local dependencies and hierarchical attention for multi-scale temporal modelling. This combination enables the encoder to learn both fine-grained and long-range temporal patterns. The encoder produces frame-level predictions for action classes and boundary classes. To refine these initial predictions, three sequential decoders \cite{zhang2023surgical,yi2021asformer,ishikawa2021alleviating,yi2022not} are employed. Each decoder shares a similar structure to the encoder but focuses on progressively refining predictions with higher accuracy and temporal smoothness.
For optimisation, we utilised the Adam optimiser along with the softmax activation function to ensure stable training. In contrast to some existing approaches \cite{yi2021asformer}, positional encoding was omitted, as the hierarchical and sliding attention mechanisms sufficiently capture temporal ordering and dependencies.
Our boundary-aware loss function is designed to enhance the model's segmentation and transition detection capabilities. It combines a weighted Dice loss, which penalises the model based on Intersection over Union (IoU), with a Gaussian Cosine Similarity loss that emphasises smooth transitions by modelling centre-to-boundary similarity. This composite loss function ensures robust learning of temporal dynamics and accurate identification of action boundaries. We split the dataset into 80-20 ratios for training testing for Jigsaw and Cholec80. We used dropout and early stopping to handle over-fitting and set the max epoch to 120. We have considered three different datasets such as SAR-RARP50 \cite{psychogyios2023sar}, JIGSAWS \cite{gao2014jhu}, Cholec80 \cite{twinanda2016endonet}, and for more details, please refer to supplementary material. 
\subsection{Evaluation metrics}
To evaluate the performance of our models, we used frame-wise accuracy along with segmental-wise overlap. The frame-wise accuracy (FWA) is calculated as $\text{FWA}=  \frac{\text{correct frames}}{\text{total frames}} $ and to asses the temporal performance, we measured the segmental scores, which include the segmental edit distance and the segmental F1 score. The Segmental F1 Score measures the temporal overlap between predicted and ground truth action segments. It is computed as the harmonic mean of precision and recall, focusing on how well the predicted segments align with the actual segments in terms of duration and timing. The segmental F1 score was assessed at overlapping thresholds of 10\%, 25\%, and 50\%, referred to as F1@10, F1@25, and F1@50, respectively. 
Determine True Positives (TP), False Positives (FP), and False Negatives (FN). A predicted segment is considered a TP if its IoU with a ground truth segment exceeds a predefined threshold \( T \). For F1@10\%, \( T = 0.10 \), For F1@25\%, \( T = 0.25 \), For F1@50\%, \( T = 0.50 \). FPs are predicted segments without a matching ground truth segment above the threshold. FNs are ground truth segments without a matching predicted segment above the threshold. Calculate Precision and Recall:
    \[
    \text{Precision} = \frac{\text{TP}_{IoU>T}}{\text{TP}_{IoU>T} + \text{FP}_{IoU>T}}
    \]
    \[
    \text{Recall} = \frac{\text{TP}_{IoU>T}}{\text{TP}_{IoU>T} + \text{FN}_{IoU>T}}
    \]
Compute Segmental F1 Score:
    \[
    \text{F1} = 2 \times \frac{\text{Precision} \times \text{Recall}}{\text{Precision} + \text{Recall}}
    \]
The Edit Score assesses the sequence of predicted action segments by comparing it to the ground truth sequence. It is calculated based on the Levenshtein distance, which quantifies the minimum number of edit operations (insertions, deletions, substitutions) required to transform the predicted sequence into the ground truth sequence. A higher Edit Score indicates a closer match between the predicted and actual sequences, reflecting the model's proficiency in capturing the correct order and transitions of actions. This metric is less sensitive to minor temporal shifts but penalises over-segmentation and out-of-order predictions. Calculate the Levenshtein distance \( D \) between the predicted and ground truth sequences. The Edit Score is then computed as $\text{Edit Score} = 1 - \frac{D}{\max(L_p, L_g)}$ where \( L_p \) and \( L_g \) are the lengths of the predicted and ground truth sequences, respectively. The Segmental F1 Score is sensitive to over-segmentation and under-segmentation errors, penalising models that divide actions into too many segments or fewer segments. In contrast, the Edit Score penalises incorrect action order and over-segmentation but is more tolerant of slight temporal misalignments. The Segmental F1 Score evaluates the spatial and temporal precision of action boundaries, whereas the Edit Score assesses the structural and sequential accuracy of the predicted action sequence.

\subsection{Results}
Our Multi-staged Boundary-aware network has three main differences from Asformer. First, we employed acausal convolution layer in the initial layers of encoders and decoders. Secondly, we used hierarchical sparse multi-head window attention. Finally, we calculated centre sampled Boundary Aware Loss leverages the strengths of both cosine similarity and Gaussian functions, making it effective for segmentation tasks. As depicted in Table~\ref{tab:sotacompare} our model outperforms other models in F1 scores and achieves comparable results in Acc and Edit with state of the art. While Accuracy and Edit Scores provide valuable insights into frame-level precision and strict sequential ordering, they are highly sensitive to slight temporal misalignment at segment boundaries. In contrast, the
F1@k metric offers greater flexibility for boundary localisation $(\pm k\%)$ making it more robust to accurately segment long actions, even with minor shifts in timing. Figure \ref{fig:merged_temporal_comparison} illustrates this qualitatively, showcasing MSBATN's superior performance in accurately defining crucial action boundaries compared to other methods like ASRF and Asformer. Table \ref{tab:gflops_params} shows MSBATN is computationally efficient in comparison with other boundary aware model like ASRF \cite{ishikawa2021alleviating}.
\begin{table*}[ht!]
\caption{Comparison of different settings Across Datasets (SAR-RARP, Jigsaw, Cholec80)}
\label{tab:comparison_settings}
\centering
\adjustbox{max width=1\textwidth}{%
\begin{tabular}{lllllllll}
\toprule
\textbf{Dataset} & \textbf{Settings} & \textbf{Acc} & \textbf{Prec} & \textbf{Rec} & \textbf{F1@10} & \textbf{F1@25} & \textbf{F1@50} & \textbf{Edit} \\ 
\midrule

\multirow{14}{*}{\rotatebox{90}{\textbf{SAR-RARP}}} 
    & CSAtt & 0.61 & 0.68 & 0.59 & 0.56 & 0.52 & 0.28 & 0.73 \\ 
    & CHAtt & 0.73 & 0.77 & 0.71 & 0.58 & 0.54 & 0.40 & 0.75 \\ 
    & CSFWNO & 0.82 & 0.89 & 0.80 & 0.77 & 0.63 & 0.49 & 0.74 \\ 
    & CSSWSO & 0.84 & 0.87 & 0.82 & 0.79 & 0.65 & 0.51 & 0.76 \\ 
    & CDWSO & 0.87 & 0.85 & 0.81 & 0.85 & 0.68 & 0.55 & 0.79 \\ 
    & CHFW & 0.84 & 0.87 & 0.81 & 0.80 & 0.66 & 0.52 & 0.76 \\ 
    & CHDSWO & 0.88 & 0.89 & 0.80 & 0.84 & 0.72 & 0.58 & 0.80 \\ 
    & ACSAtt & 0.65 & 0.67 & 0.61 & 0.57 & 0.57 & 0.43 & 0.74 \\ 
    & ACHAtt & 0.68 & 0.65 & 0.60 & 0.64 & 0.60 & 0.47 & 0.80 \\ 
    & ACSFWNO & 0.87 & 0.89 & 0.80 & 0.83 & 0.69 & 0.56 & 0.79 \\ 
    & ACDWSO & 0.88 & 0.91 & 0.84 & 0.84 & 0.71 & 0.58 & 0.81 \\ 
    & ACHFW & 0.88 & 0.90 & 0.87 & 0.82 & 0.71 & 0.59 & 0.81 \\ 
    & ACHSSWO & 0.89 & \textbf{0.92} & \textbf{0.88} & \textbf{0.85} & 0.72 & 0.60 & 0.82 \\ 
    & ACHDSWO & \textbf{0.90} & 0.86 & 0.87 & 0.87 & \textbf{0.75} & \textbf{0.64} & \textbf{0.84} \\ 
\midrule

\multirow{14}{*}{\rotatebox{90}
{\textbf{Cholec80}}} 
    & CSAtt & 0.631 & 0.712 & 0.621 & 0.601 & 0.585 & 0.472 & 0.732 \\ 
    & CHAtt & 0.748 & 0.799 & 0.728 & 0.718 & 0.604 & 0.503 & 0.751 \\ 
    & CSFWNO & 0.84 & 0.821 & 0.83 & 0.81 & 0.695 & 0.583 & 0.762 \\ 
    & CSSWSO & 0.856 & 0.837 & 0.846 & 0.826 & 0.713 & 0.604 & 0.78 \\ 
    & CDWSO & 0.89 & 0.87 & 0.86 & 0.86 & 0.77 & 0.64 & 0.84 \\ 
    & CHFW & 0.864 & 0.845 & 0.854 & 0.834 & 0.722 & 0.614 & 0.789 \\ 
    & CHDSWO & 0.878 & 0.859 & 0.868 & 0.848 & 0.738 & 0.631 & 0.805 \\ 
    & ACSAtt & 0.686 & 0.767 & 0.676 & 0.656 & 0.547 & 0.542 & 0.741 \\ 
    & ACHAtt & 0.694 & 0.754 & 0.684 & 0.764 & 0.656 & 0.553 & 0.773 \\ 
    & ACSFWNO & 0.89 & 0.871 & 0.88 & 0.86 & 0.752 & 0.648 & 0.819 \\ 
    & ACDWSO & 0.902 & 0.883 & 0.892 & 0.872 & 0.765 & 0.664 & 0.832 \\ 
    & ACSSWSO & 0.91 & 0.891 & 0.9 & 0.88 & 0.774 & 0.675 & 0.841 \\ 
    & ACHFW & 0.906 & 0.887 & 0.896 & 0.876 & 0.77 & 0.67 & 0.837 \\ 
    & ACHDSWO & \textbf{0.918} & \textbf{0.899} &\textbf{ 0.908} & \textbf{0.888} & \textbf{0.783} & \textbf{0.686} & \textbf{0.85} \\ 

\midrule

\multirow{14}{*}{\rotatebox{90}
{\textbf{JIGSAW}}} 
    & CSAtt & 0.742 & 0.823 & 0.732 & 0.712 & 0.598 & 0.488 & 0.705 \\ 
    & CHAtt & 0.759 & 0.84 & 0.79 & 0.729 & 0.617 & 0.509 & 0.748 \\ 
    & CSFWNO & 0.851 & 0.832 & 0.841 & 0.821 & 0.708 & 0.599 & 0.775 \\ 
    & CSSWSO & 0.867 & 0.848 & 0.857 & 0.837 & 0.726 & 0.620 & 0.793 \\ 
    & CDWSO & 0.89 & 0.88 & 0.887 & 0.86 & 0.76 & 0.638 & 0.809 \\ 
    & CHFW & 0.875 & 0.856 & 0.865 & 0.845 & 0.735 & 0.630 & 0.802 \\ 
    & CHDSWO & 0.889 & 0.87 & 0.879 & 0.859 & 0.751 & 0.647 & 0.818 \\ 
    & ACSAtt & 0.797 & 0.867 & 0.687 & 0.867 & 0.76 & 0.658 & 0.827 \\ 
    & ACHAtt & 0.905 & 0.886 & 0.895 & 0.875 & 0.769 & 0.669 & 0.836 \\ 
    & ACSFWNO & 0.901 & 0.882 & 0.891 & 0.871 & 0.765 & 0.664 & 0.832 \\ 
    & ACDWSO & 0.913 & 0.894 & 0.903 & 0.883 & 0.778 & 0.680 & 0.845 \\ 
    & ACHFW & 0.917 & 0.898 & 0.907 & 0.887 & 0.783 & 0.686 & 0.85 \\ 
    & ACHFSWO & \textbf{0.939} &\textbf{ 0.94} & \textbf{0.929} & \textbf{0.919} &\textbf{ 0.796} & \textbf{0.723} &\textbf{ 0.893} \\ 
    & ACHDSWO & 0.921 & 0.902 & 0.911 & 0.891 & 0.787 & 0.691 & 0.854 \\

\bottomrule
\end{tabular}%
}
\end{table*}

\subsection{Ablation studies}
We conducted an ablation study to demonstrate the effectiveness of the components in our architecture. Our configuration, which includes Acausal Convolution with Hierarchical Dual Sliding with Overlap, as shown in Table \ref{tab:comparison_settings}, outperforms other settings in long action segmentation. Additionally, we evaluated the computational complexity, as detailed in Table \ref{tab:gflops_params}. Furthermore, we compared the effectiveness of the loss functions, as presented in Table \ref{tab:comparison_loss}.

\subsubsection{Effect of Convolution and sliding window attention and overlapping }
We assessed the effectiveness of convolutions, along with various attention settings during training. As illustrated in Table \ref{tab:comparison_settings}, the causal settings include:
- Causal Self-Attention (CSAtt) 
- Causal Hierarchical Attention (CHAtt)
- Causal Fixed Sliding Window with No Overlap (CFSWNO)
- Causal Single Sliding Window with Overlap (CSSWO)
- Causal Dual Sliding Window with Overlap (CDWSO)
- Causal Hierarchical Fixed Window with Overlap (CHFWO)
- Causal Hierarchical Dual Sliding Window with Overlap (CHDSWO)
In addition, we incorporated acausal convolutions, which include:
- Acausal Self-Attention (ACSAtt)
- Acausal Hierarchical Attention (ACHAtt)
- Acausal Fixed Sliding Window with No Overlap (ACFSWNO)
- Acausal Single Sliding Window with Overlap (ACSSWO)
- Acausal Dual Sliding Window with Overlap (ACDWSO)
- Acausal Hierarchical Fixed Window with Overlap (ACHFWO)
- Acausal Hierarchical Dual Sliding Window with Overlap (ACHDSWO)

Table \ref{tab:comparison_settings} highlights the effectiveness of proposed hierarchical sliding window attention with causal and acausal convolution. Settings with acausal convolution performed better than causal convolution, as causal convolution restricts information flow to the present and past, ensuring that no future data influences the current output. In contrast, acausal convolution permits the output to depend on past, present, and future input values, which is useful during training. We also observe that self-attention and fixed hierarchical attention performed poorly. Vanilla self-attention computes attention scores between all token pairs, leading to a quadratic complexity. Such complexity becomes prohibitive for long sequences, limiting scalability. Hierarchical attention processes input sequences at multiple hierarchical levels with predetermined structures, which may not adapt well to varying local contexts within the data. Hierarchical Sliding Window Sparse Attention focuses on local neighbourhoods dynamically, capturing essential local patterns and structures. Stacking layers of such windowed attention results in a large receptive field, where the top layers have access to all input locations and can build representations that incorporate information across the entire input. Single-Sliding Window Attention performs well on the Jigsaw dataset used in surgical training, where samples are shorter and backgrounds are simpler than in real-world datasets like SAR-RARP and Cholec80. In contrast, Dual-Sliding Window Attention performs better in SAR-RARP and Cholec80, which feature complex and dynamic actions. Single-Sliding Window Attention captures short-range dependencies effectively but may struggle with long-range connections, leading to a fragmented understanding in more complex videos. Meanwhile, Dual-Sliding Window Attention is better at capturing both short- and long-range dependencies, making it suitable for handling diverse action durations and transitions.
 \begin{table*}[h!]
\centering
\caption{Performance comparison between Start-End Gaussian cosine and BCE Loss }
\label{tab:comparison_loss}
\begin{tabular}{lcccccccccc}
\toprule
\textbf{Settings} & \multicolumn{4}{c}{\textbf{Start-End Gaussian Cosine Loss}} & \multicolumn{4}{c}{\textbf{Binary Cross Entropy Loss}} \\
\cmidrule(lr){2-5} \cmidrule(lr){6-9}
 & \textbf{F1@10} & \textbf{F1@25} & \textbf{F1@50} & \textbf{Edit} & \textbf{F1@10} & \textbf{F1@25} & \textbf{F1@50} & \textbf{Edit (Bin)} \\
\midrule
ACDWSO   & 0.85 & 0.71 & 0.53 & 0.81 & 0.76 & 0.6  & 0.58 & 0.78 \\
ACHFW    & 0.83 & 0.7  & 0.55 & 0.78 & 0.7  & 0.51 & 0.45 & 0.72 \\
ACSSWSO  & 0.86 & 0.72 & 0.6  & 0.82 & 0.73 & 0.62 & 0.55 & 0.79 \\
ACHDSWO  & 0.87 & 0.74 & 0.63 & 0.84 & 0.77 & 0.64 & 0.53 & 0.8  \\
\bottomrule
\end{tabular}
\end{table*}
\begin{table}[ht!]
  \caption{Impact of HSWA Components (DSWA and HTA) and GFLOPs and parameters.}
  \label{tab:gflops_params}
  \centering
  \begin{tabular}{ p{10cm}  c  c  c  c }
    \toprule
    \textbf{Model} & \textbf{Acc} & \textbf{F1@10} & \textbf{GFLOPs} & \textbf{Parameters} \\
    \midrule
    \raggedright Transformer~\cite{vaswani2017attention}
      & 65.3 & 57.8   & 204.47  & 20.5181\,M  \\ 
    \raggedright ASRF~\cite{ishikawa2021alleviating}
      & 87.6  &80.3  & 136.42  & 13.6686\,M  \\
    \raggedright NETE~\cite{yi2022not}
      & 87.9  & 79.4   & 131.10  & 13.1454\,M  \\
      \raggedright Asformer~\cite{yi2021asformer}
      & 88.5  & 84.5   & 101.03  & 10.0027\,M  \\
      \raggedright \textbf{DSWA}
      & 86.8  & 82.6   & 101.31  & 7.9750\,M  \\
      \raggedright \textbf{HTA}
      & 84.4  & 76.2   &  76.78 & 7.8594\,M  \\
    \raggedright \textbf{HSWA(DSWA+HTA)}
      & \textbf{88.7 } & \textbf{84.5}   & \textbf{104.53}  & \textbf{11.9450\,M}  \\
    \bottomrule
  \end{tabular}
\end{table}

\subsubsection{Effect of Start-End Boundary Aware Loss Function}

Traditional loss functions, such as Binary Cross-Entropy (BCE), predominantly focus on frame-wise classification, which can result in either over-segmentation or under-segmentation. This approach has inherent limitations, particularly when addressing complex or ambiguous boundaries. Additionally, binary boundary classification may lead to instability during training due to substantial variations in loss values both within and between classes.
The results of the proposed boundary-aware loss function show that our proposed approach effectively handled these challenges by capitalising on the distribution of similarity scores between frames generated by decoders. Table \ref{tab:comparison_loss} demonstrates the effectiveness of our method, which incorporates acausal convolution and sliding window attention settings, delivering improved results across the five metrics outlined in the previous section compared to the binary-boundary-based approach.

\section{Conclusion and Feature Work}\label{sec5}

Segmenting actions in lengthy surgical settings is challenging due to the complex interactions between tools and organs. There are no established standards for optimal performance, as most assessments rely on frame-wise recognition, which has now shifted to segment-wise understanding. In this article, we present a TCN-transformer-based framework with hierarchical sliding window attention that operates with linear computational costs.

We found that selecting the right attention module can be difficult, and neighbouring frames often lead to ambiguous results, causing over- and under-segmentation in frame-wise segmentation. Segment-wise refinement is more effective in improving the understanding of surgical activities. Additionally, binary label boundaries are often inadequate because the start, centre, and end of actions vary by individual expertise, leading to poor performance. Our proposed framework addresses these issues with hierarchical sliding window attention for global-local context and includes a novel Boundary Aware Segmentation Loss to better recognise central and boundary points. Our boundary-aware segmentation approach outperforms traditional binary detection methods.
Experimental results show that our model matches or exceeds the performance of current state-of-the-art methods across various metrics, ensuring robustness in complex surgical environments.

These results open new interesting research directions, in particular, exploring anticipation tasks where decoders predict the next segment while accounting for uncertainty in a goal-driven manner. Furthermore, integrating instructional prompts to enhance and refine the learning process represents another promising avenue for future investigation.










\printcredits

\bibliographystyle{cas-model2-names}

\subsection*{Declaration of generative AI and AI-assisted technologies in the writing process}
During the preparation of this work, the author(s) used Grammarly and ChatGPT in order to refine the writing, check grammar, and improve overall readability. After using these tools, the author(s) reviewed and edited the content as needed and take(s) full responsibility for the content of the publication.

\bibliography{example}

\begin{thebibliography}{38}
\expandafter\ifx\csname natexlab\endcsname\relax\def\natexlab#1{#1}\fi
\providecommand{\url}[1]{\texttt{#1}}
\providecommand{\href}[2]{#2}
\providecommand{\path}[1]{#1}
\providecommand{\DOIprefix}{doi:}
\providecommand{\ArXivprefix}{arXiv:}
\providecommand{\URLprefix}{URL: }
\providecommand{\Pubmedprefix}{pmid:}
\providecommand{\doi}[1]{\href{http://dx.doi.org/#1}{\path{#1}}}
\providecommand{\Pubmed}[1]{\href{pmid:#1}{\path{#1}}}
\providecommand{\bibinfo}[2]{#2}
\ifx\xfnm\relax \def\xfnm[#1]{\unskip,\space#1}\fi
\bibitem[{Beltagy et~al.(2020)Beltagy, Peters and Cohan}]{beltagy2020longformer}
\bibinfo{author}{Beltagy, I.}, \bibinfo{author}{Peters, M.E.}, \bibinfo{author}{Cohan, A.}, \bibinfo{year}{2020}.
\newblock \bibinfo{title}{Longformer: The long-document transformer}.
\newblock \bibinfo{journal}{arXiv preprint arXiv:2004.05150} .
\bibitem[{Chadebecq et~al.(2023)Chadebecq, Lovat and Stoyanov}]{chadebecq2023artificial}
\bibinfo{author}{Chadebecq, F.}, \bibinfo{author}{Lovat, L.B.}, \bibinfo{author}{Stoyanov, D.}, \bibinfo{year}{2023}.
\newblock \bibinfo{title}{Artificial intelligence and automation in endoscopy and surgery}.
\newblock \bibinfo{journal}{Nature Reviews Gastroenterology \& Hepatology} \bibinfo{volume}{20}, \bibinfo{pages}{171--182}.
\bibitem[{Chao et~al.(2018)Chao, Vijayanarasimhan, Seybold, Ross, Deng and Sukthankar}]{chao2018rethinking}
\bibinfo{author}{Chao, Y.W.}, \bibinfo{author}{Vijayanarasimhan, S.}, \bibinfo{author}{Seybold, B.}, \bibinfo{author}{Ross, D.A.}, \bibinfo{author}{Deng, J.}, \bibinfo{author}{Sukthankar, R.}, \bibinfo{year}{2018}.
\newblock \bibinfo{title}{Rethinking the faster r-cnn architecture for temporal action localization}, in: \bibinfo{booktitle}{Proceedings of the IEEE conference on computer vision and pattern recognition}, pp. \bibinfo{pages}{1130--1139}.
\bibitem[{Czempiel et~al.(2020)Czempiel, Paschali, Keicher, Simson, Feussner, Kim and Navab}]{czempiel2020tecno}
\bibinfo{author}{Czempiel, T.}, \bibinfo{author}{Paschali, M.}, \bibinfo{author}{Keicher, M.}, \bibinfo{author}{Simson, W.}, \bibinfo{author}{Feussner, H.}, \bibinfo{author}{Kim, S.T.}, \bibinfo{author}{Navab, N.}, \bibinfo{year}{2020}.
\newblock \bibinfo{title}{Tecno: Surgical phase recognition with multi-stage temporal convolutional networks}, in: \bibinfo{booktitle}{Medical Image Computing and Computer Assisted Intervention--MICCAI 2020: 23rd International Conference, Lima, Peru, October 4--8, 2020, Proceedings, Part III 23}, \bibinfo{organization}{Springer}. pp. \bibinfo{pages}{343--352}.
\bibitem[{Ding and Li(2022)}]{ding2022exploring}
\bibinfo{author}{Ding, X.}, \bibinfo{author}{Li, X.}, \bibinfo{year}{2022}.
\newblock \bibinfo{title}{Exploring segment-level semantics for online phase recognition from surgical videos}.
\newblock \bibinfo{journal}{IEEE Transactions on Medical Imaging} \bibinfo{volume}{41}, \bibinfo{pages}{3309--3319}.
\newblock \DOIprefix\doi{10.1109/TMI.2022.3182995}.
\bibitem[{Fard et~al.(2017)Fard, Ameri, Chinnam and Ellis}]{fard2017soft}
\bibinfo{author}{Fard, M.J.}, \bibinfo{author}{Ameri, S.}, \bibinfo{author}{Chinnam, R.B.}, \bibinfo{author}{Ellis, R.D.}, \bibinfo{year}{2017}.
\newblock \bibinfo{title}{Soft boundary approach for unsupervised gesture segmentation in robotic-assisted surgery}.
\newblock \bibinfo{journal}{IEEE Robotics and Automation Letters} \bibinfo{volume}{2}, \bibinfo{pages}{171--178}.
\newblock \DOIprefix\doi{10.1109/LRA.2016.2585303}.
\bibitem[{Farha and Gall(2019)}]{farha2019ms}
\bibinfo{author}{Farha, Y.A.}, \bibinfo{author}{Gall, J.}, \bibinfo{year}{2019}.
\newblock \bibinfo{title}{Ms-tcn: Multi-stage temporal convolutional network for action segmentation}, in: \bibinfo{booktitle}{Proceedings of the IEEE/CVF conference on computer vision and pattern recognition}, pp. \bibinfo{pages}{3575--3584}.
\bibitem[{Gao et~al.(2014)Gao, Vedula, Reiley, Ahmidi, Varadarajan, Lin, Tao, Zappella, B{\'e}jar, Yuh et~al.}]{gao2014jhu}
\bibinfo{author}{Gao, Y.}, \bibinfo{author}{Vedula, S.S.}, \bibinfo{author}{Reiley, C.E.}, \bibinfo{author}{Ahmidi, N.}, \bibinfo{author}{Varadarajan, B.}, \bibinfo{author}{Lin, H.C.}, \bibinfo{author}{Tao, L.}, \bibinfo{author}{Zappella, L.}, \bibinfo{author}{B{\'e}jar, B.}, \bibinfo{author}{Yuh, D.D.}, et~al., \bibinfo{year}{2014}.
\newblock \bibinfo{title}{Jhu-isi gesture and skill assessment working set (jigsaws): A surgical activity dataset for human motion modeling}, in: \bibinfo{booktitle}{MICCAI workshop: M2cai}, p.~\bibinfo{pages}{3}.
\bibitem[{Goldbraikh et~al.(2023)Goldbraikh, Shubi, Rubin, Pugh and Laufer}]{goldbraikh2023kinematic}
\bibinfo{author}{Goldbraikh, A.}, \bibinfo{author}{Shubi, O.}, \bibinfo{author}{Rubin, O.}, \bibinfo{author}{Pugh, C.M.}, \bibinfo{author}{Laufer, S.}, \bibinfo{year}{2023}.
\newblock \bibinfo{title}{Kinematic data-based action segmentation for surgical applications}.
\newblock \bibinfo{journal}{arXiv preprint arXiv:2303.07814} .
\bibitem[{Huang et~al.(2022)Huang, Jin, Gong and Liu}]{huang2022video}
\bibinfo{author}{Huang, J.}, \bibinfo{author}{Jin, H.}, \bibinfo{author}{Gong, S.}, \bibinfo{author}{Liu, Y.}, \bibinfo{year}{2022}.
\newblock \bibinfo{title}{Video activity localisation with uncertainties in temporal boundary}, in: \bibinfo{booktitle}{European Conference on Computer Vision}, \bibinfo{organization}{Springer}. pp. \bibinfo{pages}{724--740}.
\bibitem[{Ishikawa et~al.(2021)Ishikawa, Kasai, Aoki and Kataoka}]{ishikawa2021alleviating}
\bibinfo{author}{Ishikawa, Y.}, \bibinfo{author}{Kasai, S.}, \bibinfo{author}{Aoki, Y.}, \bibinfo{author}{Kataoka, H.}, \bibinfo{year}{2021}.
\newblock \bibinfo{title}{Alleviating over-segmentation errors by detecting action boundaries}, in: \bibinfo{booktitle}{Proceedings of the IEEE/CVF winter conference on applications of computer vision}, pp. \bibinfo{pages}{2322--2331}.
\bibitem[{Jin et~al.(2020)Jin, Li, Dou, Chen, Qin, Fu and Heng}]{jin2020multi}
\bibinfo{author}{Jin, Y.}, \bibinfo{author}{Li, H.}, \bibinfo{author}{Dou, Q.}, \bibinfo{author}{Chen, H.}, \bibinfo{author}{Qin, J.}, \bibinfo{author}{Fu, C.W.}, \bibinfo{author}{Heng, P.A.}, \bibinfo{year}{2020}.
\newblock \bibinfo{title}{Multi-task recurrent convolutional network with correlation loss for surgical video analysis}.
\newblock \bibinfo{journal}{Medical image analysis} \bibinfo{volume}{59}, \bibinfo{pages}{101572}.
\bibitem[{Jin et~al.(2022)Jin, Yu, Chen, Zhao, Heng and Stoyanov}]{jin2022exploring}
\bibinfo{author}{Jin, Y.}, \bibinfo{author}{Yu, Y.}, \bibinfo{author}{Chen, C.}, \bibinfo{author}{Zhao, Z.}, \bibinfo{author}{Heng, P.A.}, \bibinfo{author}{Stoyanov, D.}, \bibinfo{year}{2022}.
\newblock \bibinfo{title}{Exploring intra-and inter-video relation for surgical semantic scene segmentation}.
\newblock \bibinfo{journal}{IEEE Transactions on Medical Imaging} \bibinfo{volume}{41}, \bibinfo{pages}{2991--3002}.
\bibitem[{Keisham et~al.(2022)Keisham, Jalali and Lee}]{keisham2022online}
\bibinfo{author}{Keisham, K.}, \bibinfo{author}{Jalali, A.}, \bibinfo{author}{Lee, M.}, \bibinfo{year}{2022}.
\newblock \bibinfo{title}{Online action proposal generation using spatio-temporal attention network}.
\newblock \bibinfo{journal}{Neural Networks} \bibinfo{volume}{153}, \bibinfo{pages}{518--529}.
\bibitem[{Kiyasseh et~al.(2023)Kiyasseh, Ma, Haque, Miles, Wagner, Donoho, Anandkumar and Hung}]{kiyasseh2023vision}
\bibinfo{author}{Kiyasseh, D.}, \bibinfo{author}{Ma, R.}, \bibinfo{author}{Haque, T.F.}, \bibinfo{author}{Miles, B.J.}, \bibinfo{author}{Wagner, C.}, \bibinfo{author}{Donoho, D.A.}, \bibinfo{author}{Anandkumar, A.}, \bibinfo{author}{Hung, A.J.}, \bibinfo{year}{2023}.
\newblock \bibinfo{title}{A vision transformer for decoding surgeon activity from surgical videos}.
\newblock \bibinfo{journal}{Nature biomedical engineering} \bibinfo{volume}{7}, \bibinfo{pages}{780--796}.
\bibitem[{Li et~al.(2020)Li, Farha, Liu, Cheng and Gall}]{li2020ms}
\bibinfo{author}{Li, S.}, \bibinfo{author}{Farha, Y.A.}, \bibinfo{author}{Liu, Y.}, \bibinfo{author}{Cheng, M.M.}, \bibinfo{author}{Gall, J.}, \bibinfo{year}{2020}.
\newblock \bibinfo{title}{Ms-tcn++: Multi-stage temporal convolutional network for action segmentation}.
\newblock \bibinfo{journal}{IEEE transactions on pattern analysis and machine intelligence} \bibinfo{volume}{45}, \bibinfo{pages}{6647--6658}.
\bibitem[{Loukas et~al.(2016)Loukas, Nikiteas, Schizas and Georgiou}]{loukas2016shot}
\bibinfo{author}{Loukas, C.}, \bibinfo{author}{Nikiteas, N.}, \bibinfo{author}{Schizas, D.}, \bibinfo{author}{Georgiou, E.}, \bibinfo{year}{2016}.
\newblock \bibinfo{title}{Shot boundary detection in endoscopic surgery videos using a variational bayesian framework}.
\newblock \bibinfo{journal}{International journal of computer assisted radiology and surgery} \bibinfo{volume}{11}, \bibinfo{pages}{1937--1949}.
\bibitem[{Mascagni et~al.(2022)Mascagni, Alapatt, Sestini, Altieri, Madani, Watanabe, Alseidi, Redan, Alfieri, Costamagna et~al.}]{mascagni2022computer}
\bibinfo{author}{Mascagni, P.}, \bibinfo{author}{Alapatt, D.}, \bibinfo{author}{Sestini, L.}, \bibinfo{author}{Altieri, M.S.}, \bibinfo{author}{Madani, A.}, \bibinfo{author}{Watanabe, Y.}, \bibinfo{author}{Alseidi, A.}, \bibinfo{author}{Redan, J.A.}, \bibinfo{author}{Alfieri, S.}, \bibinfo{author}{Costamagna, G.}, et~al., \bibinfo{year}{2022}.
\newblock \bibinfo{title}{Computer vision in surgery: from potential to clinical value}.
\newblock \bibinfo{journal}{npj Digital Medicine} \bibinfo{volume}{5}, \bibinfo{pages}{163}.
\bibitem[{Meireles et~al.(2021)Meireles, Rosman, Altieri, Carin, Hager, Madani, Padoy, Pugh, Sylla, Ward et~al.}]{meireles2021sages}
\bibinfo{author}{Meireles, O.R.}, \bibinfo{author}{Rosman, G.}, \bibinfo{author}{Altieri, M.S.}, \bibinfo{author}{Carin, L.}, \bibinfo{author}{Hager, G.}, \bibinfo{author}{Madani, A.}, \bibinfo{author}{Padoy, N.}, \bibinfo{author}{Pugh, C.M.}, \bibinfo{author}{Sylla, P.}, \bibinfo{author}{Ward, T.M.}, et~al., \bibinfo{year}{2021}.
\newblock \bibinfo{title}{Sages consensus recommendations on an annotation framework for surgical video}.
\newblock \bibinfo{journal}{Surgical endoscopy} \bibinfo{volume}{35}, \bibinfo{pages}{4918--4929}.
\bibitem[{Mondal et~al.(2019)Mondal, Sathish and Sheet}]{mondal2019multitask}
\bibinfo{author}{Mondal, S.S.}, \bibinfo{author}{Sathish, R.}, \bibinfo{author}{Sheet, D.}, \bibinfo{year}{2019}.
\newblock \bibinfo{title}{Multitask learning of temporal connectionism in convolutional networks using a joint distribution loss function to simultaneously identify tools and phase in surgical videos}.
\newblock \bibinfo{journal}{arXiv preprint arXiv:1905.08315} .
\bibitem[{Namazi et~al.(2019)Namazi, Sankaranarayanan and Devarajan}]{namazi2019attention}
\bibinfo{author}{Namazi, B.}, \bibinfo{author}{Sankaranarayanan, G.}, \bibinfo{author}{Devarajan, V.}, \bibinfo{year}{2019}.
\newblock \bibinfo{title}{Attention-based surgical phase boundaries detection in laparoscopic videos}, in: \bibinfo{booktitle}{2019 International Conference on Computational Science and Computational Intelligence (CSCI)}, pp. \bibinfo{pages}{577--583}.
\newblock \DOIprefix\doi{10.1109/CSCI49370.2019.00109}.
\bibitem[{Neil et~al.(2016)Neil, Pfeiffer and Liu}]{neil2016phased}
\bibinfo{author}{Neil, D.}, \bibinfo{author}{Pfeiffer, M.}, \bibinfo{author}{Liu, S.C.}, \bibinfo{year}{2016}.
\newblock \bibinfo{title}{Phased lstm: Accelerating recurrent network training for long or event-based sequences}.
\newblock \bibinfo{journal}{Advances in neural information processing systems} \bibinfo{volume}{29}.
\bibitem[{Nwoye et~al.(2023)Nwoye, Alapatt, Yu, Vardazaryan, Xia, Zhao, Xia, Jia, Yang, Wang et~al.}]{nwoye2023cholectriplet2021}
\bibinfo{author}{Nwoye, C.I.}, \bibinfo{author}{Alapatt, D.}, \bibinfo{author}{Yu, T.}, \bibinfo{author}{Vardazaryan, A.}, \bibinfo{author}{Xia, F.}, \bibinfo{author}{Zhao, Z.}, \bibinfo{author}{Xia, T.}, \bibinfo{author}{Jia, F.}, \bibinfo{author}{Yang, Y.}, \bibinfo{author}{Wang, H.}, et~al., \bibinfo{year}{2023}.
\newblock \bibinfo{title}{Cholectriplet2021: A benchmark challenge for surgical action triplet recognition}.
\newblock \bibinfo{journal}{Medical Image Analysis} \bibinfo{volume}{86}, \bibinfo{pages}{102803}.
\bibitem[{Nwoye et~al.(2022)Nwoye, Yu, Gonzalez, Seeliger, Mascagni, Mutter, Marescaux and Padoy}]{nwoye2022rendezvous}
\bibinfo{author}{Nwoye, C.I.}, \bibinfo{author}{Yu, T.}, \bibinfo{author}{Gonzalez, C.}, \bibinfo{author}{Seeliger, B.}, \bibinfo{author}{Mascagni, P.}, \bibinfo{author}{Mutter, D.}, \bibinfo{author}{Marescaux, J.}, \bibinfo{author}{Padoy, N.}, \bibinfo{year}{2022}.
\newblock \bibinfo{title}{Rendezvous: Attention mechanisms for the recognition of surgical action triplets in endoscopic videos}.
\newblock \bibinfo{journal}{Medical Image Analysis} \bibinfo{volume}{78}, \bibinfo{pages}{102433}.
\bibitem[{Oquab et~al.(2023)Oquab, Darcet, Moutakanni, Vo, Szafraniec, Khalidov, Fernandez, Haziza, Massa, El-Nouby et~al.}]{oquab2023dinov2}
\bibinfo{author}{Oquab, M.}, \bibinfo{author}{Darcet, T.}, \bibinfo{author}{Moutakanni, T.}, \bibinfo{author}{Vo, H.}, \bibinfo{author}{Szafraniec, M.}, \bibinfo{author}{Khalidov, V.}, \bibinfo{author}{Fernandez, P.}, \bibinfo{author}{Haziza, D.}, \bibinfo{author}{Massa, F.}, \bibinfo{author}{El-Nouby, A.}, et~al., \bibinfo{year}{2023}.
\newblock \bibinfo{title}{Dinov2: Learning robust visual features without supervision}.
\newblock \bibinfo{journal}{arXiv preprint arXiv:2304.07193} .
\bibitem[{Psychogyios et~al.(2023)Psychogyios, Colleoni, Van~Amsterdam, Li, Huang, Li, Jia, Zou, Wang, Liu et~al.}]{psychogyios2023sar}
\bibinfo{author}{Psychogyios, D.}, \bibinfo{author}{Colleoni, E.}, \bibinfo{author}{Van~Amsterdam, B.}, \bibinfo{author}{Li, C.Y.}, \bibinfo{author}{Huang, S.Y.}, \bibinfo{author}{Li, Y.}, \bibinfo{author}{Jia, F.}, \bibinfo{author}{Zou, B.}, \bibinfo{author}{Wang, G.}, \bibinfo{author}{Liu, Y.}, et~al., \bibinfo{year}{2023}.
\newblock \bibinfo{title}{Sar-rarp50: Segmentation of surgical instrumentation and action recognition on robot-assisted radical prostatectomy challenge}.
\newblock \bibinfo{journal}{arXiv preprint arXiv:2401.00496} .
\bibitem[{Ramesh et~al.(2021)Ramesh, Dall’Alba, Gonzalez, Yu, Mascagni, Mutter, Marescaux, Fiorini and Padoy}]{ramesh2021multi}
\bibinfo{author}{Ramesh, S.}, \bibinfo{author}{Dall’Alba, D.}, \bibinfo{author}{Gonzalez, C.}, \bibinfo{author}{Yu, T.}, \bibinfo{author}{Mascagni, P.}, \bibinfo{author}{Mutter, D.}, \bibinfo{author}{Marescaux, J.}, \bibinfo{author}{Fiorini, P.}, \bibinfo{author}{Padoy, N.}, \bibinfo{year}{2021}.
\newblock \bibinfo{title}{Multi-task temporal convolutional networks for joint recognition of surgical phases and steps in gastric bypass procedures}.
\newblock \bibinfo{journal}{International journal of computer assisted radiology and surgery} \bibinfo{volume}{16}, \bibinfo{pages}{1111--1119}.
\bibitem[{Shi et~al.(2023)Shi, Zhong, Cao, Ma, Li and Tao}]{shi2023tridet}
\bibinfo{author}{Shi, D.}, \bibinfo{author}{Zhong, Y.}, \bibinfo{author}{Cao, Q.}, \bibinfo{author}{Ma, L.}, \bibinfo{author}{Li, J.}, \bibinfo{author}{Tao, D.}, \bibinfo{year}{2023}.
\newblock \bibinfo{title}{Tridet: Temporal action detection with relative boundary modeling}, in: \bibinfo{booktitle}{Proceedings of the IEEE/CVF Conference on Computer Vision and Pattern Recognition}, pp. \bibinfo{pages}{18857--18866}.
\bibitem[{Twinanda(2017)}]{twinanda2017vision}
\bibinfo{author}{Twinanda, A.P.}, \bibinfo{year}{2017}.
\newblock \bibinfo{title}{Vision-based approaches for surgical activity recognition using laparoscopic and RBGD videos}.
\newblock Ph.D. thesis. Strasbourg.
\bibitem[{Twinanda et~al.(2016)Twinanda, Shehata, Mutter, Marescaux, De~Mathelin and Padoy}]{twinanda2016endonet}
\bibinfo{author}{Twinanda, A.P.}, \bibinfo{author}{Shehata, S.}, \bibinfo{author}{Mutter, D.}, \bibinfo{author}{Marescaux, J.}, \bibinfo{author}{De~Mathelin, M.}, \bibinfo{author}{Padoy, N.}, \bibinfo{year}{2016}.
\newblock \bibinfo{title}{Endonet: a deep architecture for recognition tasks on laparoscopic videos}.
\newblock \bibinfo{journal}{IEEE transactions on medical imaging} \bibinfo{volume}{36}, \bibinfo{pages}{86--97}.
\bibitem[{Twinanda et~al.(2018)Twinanda, Yengera, Mutter, Marescaux and Padoy}]{twinanda2018rsdnet}
\bibinfo{author}{Twinanda, A.P.}, \bibinfo{author}{Yengera, G.}, \bibinfo{author}{Mutter, D.}, \bibinfo{author}{Marescaux, J.}, \bibinfo{author}{Padoy, N.}, \bibinfo{year}{2018}.
\newblock \bibinfo{title}{Rsdnet: Learning to predict remaining surgery duration from laparoscopic videos without manual annotations}.
\newblock \bibinfo{journal}{IEEE transactions on medical imaging} \bibinfo{volume}{38}, \bibinfo{pages}{1069--1078}.
\bibitem[{Van~Amsterdam et~al.(2022)Van~Amsterdam, Funke, Edwards, Speidel, Collins, Sridhar, Kelly, Clarkson and Stoyanov}]{van2022gesture}
\bibinfo{author}{Van~Amsterdam, B.}, \bibinfo{author}{Funke, I.}, \bibinfo{author}{Edwards, E.}, \bibinfo{author}{Speidel, S.}, \bibinfo{author}{Collins, J.}, \bibinfo{author}{Sridhar, A.}, \bibinfo{author}{Kelly, J.}, \bibinfo{author}{Clarkson, M.J.}, \bibinfo{author}{Stoyanov, D.}, \bibinfo{year}{2022}.
\newblock \bibinfo{title}{Gesture recognition in robotic surgery with multimodal attention}.
\newblock \bibinfo{journal}{IEEE Transactions on Medical Imaging} \bibinfo{volume}{41}, \bibinfo{pages}{1677--1687}.
\bibitem[{Vaswani et~al.(2017)Vaswani, Shazeer, Parmar, Uszkoreit, Jones, Gomez, Kaiser and Polosukhin}]{vaswani2017attention}
\bibinfo{author}{Vaswani, A.}, \bibinfo{author}{Shazeer, N.}, \bibinfo{author}{Parmar, N.}, \bibinfo{author}{Uszkoreit, J.}, \bibinfo{author}{Jones, L.}, \bibinfo{author}{Gomez, A.N.}, \bibinfo{author}{Kaiser, {\L}.}, \bibinfo{author}{Polosukhin, I.}, \bibinfo{year}{2017}.
\newblock \bibinfo{title}{Attention is all you need}.
\newblock \bibinfo{journal}{Advances in neural information processing systems} \bibinfo{volume}{30}.
\bibitem[{Yang et~al.(2024)Yang, Luo, Wang and Chen}]{yang2024surgformer}
\bibinfo{author}{Yang, S.}, \bibinfo{author}{Luo, L.}, \bibinfo{author}{Wang, Q.}, \bibinfo{author}{Chen, H.}, \bibinfo{year}{2024}.
\newblock \bibinfo{title}{Surgformer: Surgical transformer with hierarchical temporal attention for surgical phase recognition}.
\newblock \bibinfo{journal}{arXiv preprint arXiv:2408.03867} .
\bibitem[{Yi et~al.(2021)Yi, Wen and Jiang}]{yi2021asformer}
\bibinfo{author}{Yi, F.}, \bibinfo{author}{Wen, H.}, \bibinfo{author}{Jiang, T.}, \bibinfo{year}{2021}.
\newblock \bibinfo{title}{Asformer: Transformer for action segmentation}.
\newblock \bibinfo{journal}{arXiv preprint arXiv:2110.08568} .
\bibitem[{Yi et~al.(2022)Yi, Yang and Jiang}]{yi2022not}
\bibinfo{author}{Yi, F.}, \bibinfo{author}{Yang, Y.}, \bibinfo{author}{Jiang, T.}, \bibinfo{year}{2022}.
\newblock \bibinfo{title}{Not end-to-end: Explore multi-stage architecture for online surgical phase recognition}, in: \bibinfo{booktitle}{Proceedings of the Asian Conference on Computer Vision}, pp. \bibinfo{pages}{2613--2628}.
\bibitem[{Zhang et~al.(2023)Zhang, Goel, Sarhan, Goel, Abukhalil, Kalesan, Stottler and Petculescu}]{zhang2023surgical}
\bibinfo{author}{Zhang, B.}, \bibinfo{author}{Goel, B.}, \bibinfo{author}{Sarhan, M.H.}, \bibinfo{author}{Goel, V.K.}, \bibinfo{author}{Abukhalil, R.}, \bibinfo{author}{Kalesan, B.}, \bibinfo{author}{Stottler, N.}, \bibinfo{author}{Petculescu, S.}, \bibinfo{year}{2023}.
\newblock \bibinfo{title}{Surgical workflow recognition with temporal convolution and transformer for action segmentation}.
\newblock \bibinfo{journal}{International Journal of Computer Assisted Radiology and Surgery} \bibinfo{volume}{18}, \bibinfo{pages}{785--794}.
\bibitem[{Zhang et~al.(2022)Zhang, Wu and Li}]{zhang2022actionformer}
\bibinfo{author}{Zhang, C.L.}, \bibinfo{author}{Wu, J.}, \bibinfo{author}{Li, Y.}, \bibinfo{year}{2022}.
\newblock \bibinfo{title}{Actionformer: Localizing moments of actions with transformers}, in: \bibinfo{booktitle}{European Conference on Computer Vision}, \bibinfo{organization}{Springer}. pp. \bibinfo{pages}{492--510}.

\end{thebibliography}



\end{document}